\documentclass[a4paper,fleqn]{cas-sc}

\usepackage[authoryear]{natbib}
\usepackage{enumitem,subfigure,appendix,lineno,eurosym}

\newcommand{\DefStat}[2]{%
  \expandafter\newcommand\csname stat-#1\endcsname{#2}%
}
\newcommand{\Stat}[1]{\csname stat-#1\endcsname}

\DefStat{totaldatapoints}{22,637}
\DefStat{totalimages}{19,750}
\DefStat{totalparticipants}{331}
\DefStat{genderwoman}{40.5\% (134)}
\DefStat{genderman}{52.3\% (173)}
\DefStat{gendernonbinary}{0.9\% (3)}
\DefStat{nlparticipants}{70.7\% (234)}
\DefStat{euparticipants}{8.8\% (29)}
\DefStat{noneuparticipants}{7.6\% (25)}
\DefStat{eduprimary}{0.3\% (1)}
\DefStat{edusecondary}{3.0\% (10)}
\DefStat{edutertiary}{39.9\% (132)}
\DefStat{edupostgraduate}{52.9\% (175)}
\DefStat{age1827}{48.6\% (161)}
\DefStat{age2837}{32.6\% (108)}
\DefStat{age3847}{11.2\% (37)}
\DefStat{age4857}{5.1\% (17)}
\DefStat{age5867}{1.8\% (6)}
\DefStat{age6877}{0.3\% (1)}
\DefStat{age7887}{0.0\% (0)}
\DefStat{age8897}{0.0\% (0)}
\DefStat{ageavg}{30}
\DefStat{agestddev}{10}
\DefStat{income00000999}{27.8\% (92)}
\DefStat{income10001999}{11.2\% (37)}
\DefStat{income20002999}{12.7\% (42)}
\DefStat{income30003999}{11.5\% (38)}
\DefStat{income40004999}{6.3\% (21)}
\DefStat{income50005999}{3.3\% (11)}
\DefStat{income00001999}{39.0\% (129)}
\DefStat{income20003999}{24.2\% (80)}
\DefStat{income400015000}{12.4\% (41)}
\DefStat{mediandurationminutes}{6.3}
\DefStat{averagedurationminutes}{42.9}
\DefStat{duration30}{93.4\% (309)}
\DefStat{duration1440}{5.4\% (18)}
\DefStat{medianintervalseconds}{3.7}
\DefStat{averageintervalseconds}{38.2}
\DefStat{intervalseconds10}{89.7\% (20,310)}
\DefStat{averageratingspersurvey}{68.4}
\DefStat{medianratingspersurvey}{100.0}
\DefStat{completedsurveys25}{77.3\% (256)}
\DefStat{completedsurveys50}{59.2\% (196)}
\DefStat{completedsurveys75}{52.6\% (174)}
\DefStat{completedsurveys100}{50.8\% (168)}
\DefStat{genderother}{6.3\% (21)}
\DefStat{otherparticipants}{13.0\% (43)}
\DefStat{eduother}{3.9\% (13)}
\DefStat{incomeother}{24.5\% (81)}
\DefStat{rps001049}{40.8\% (135)}
\DefStat{rps050099}{8.5\% (28)}
\DefStat{rps100149}{49.8\% (165)}
\DefStat{rps150999}{0.9\% (3)}

\newcommand\filename[1]{{\tt #1}} %

\newcommand\apifn[1]{\texttt{#1}}
\newcommand\apiparam[1]{\texttt{#1}}

\newcommand{\prn}[1]{\ensuremath{\left({#1}\right)}}
\newcommand{\func}[2]{\ensuremath{{#1}\!\prn{#2}}}
\newcommand{\R}[1]{\func{R}{#1}}
\newcommand{\B}[1]{\func{B}{#1}}
\newcommand{\C}[1]{\func{C}{#1}}

\begin{document}
\let\WriteBookmarks\relax
\def\floatpagepagefraction{1}
\def\textpagefraction{.001}

\shorttitle{Perceptions of street views}

\shortauthors{M. Danish et~al.}

\title [mode = title]{A citizen science toolkit to collect human perceptions of urban environments using open street view images}
\author{Matthew Danish}[type=editor,
                        auid=000,bioid=1,
                        orcid=0000-0002-7186-387X]

\cormark[1]

\ead{m.r.danish@uu.nl}

\credit{Conceptualization, Writing, Software, Methodology, Visualization}

\affiliation{organization={Utrecht University},
    addressline={Princetonlaan 8a},
    city={Utrecht},
    citysep={}, %
    postcode={3584 CB}, 
    country={The Netherlands}}

\author{S.M. Labib}

\credit{Conceptualization, Writing, Funding acquisition, Methodology}

\author{Britta Ricker}

\credit{Conceptualization, Writing, Funding acquisition, Methodology, Visualization}

\author{Marco Helbich}

\credit{Conceptualization, Writing, Funding acquisition, Project administration, Methodology}

\cortext[cor1]{Corresponding author}

\begin{abstract}
Street View Imagery (SVI) is a valuable data source for studies (e.g., environmental assessments, green space identification or land cover classification). While commercial SVI is available, such providers commonly restrict copying or reuse in ways necessary for research. Open SVI datasets are readily available from less restrictive sources, such as Mapillary, but due to the heterogeneity of the images, these require substantial preprocessing, filtering, and careful quality checks.
We present a method for automated downloading, processing, cropping, and filtering open SVI, to be used in a survey of human perceptions of the streets portrayed in these images. We demonstrate our open-source reusable SVI preparation and smartphone-friendly perception-survey software with Amsterdam (Netherlands) as the case study. Using a citizen science approach, we collected from \Stat{totalparticipants} people \Stat{totaldatapoints} ratings about their perceptions for various criteria. We have published our software in a public repository for future re-use and reproducibility.
\end{abstract}

\begin{highlights}
\item Studies using commercial street view imagery have proliferated despite licensing terms.
\item We built a workflow and webapp to collect perceptions of open street view imagery.
\item The webapp presents a simple and consistent interface with a swipe-to-rate UI.
\item Our data preparation methods and mobile-friendly survey are open, FAIR and reusable.
\item Anyone may easily clone, modify \& deploy this perception survey in any desired place.
\end{highlights}


\maketitle

\section{Introduction}

`Would you feel safe in this place? Does it look pleasant? Does it feel walkable?' 
These are some examples of questions that we can ask people about their view of a city from the street level. The answers will vary from person to person, and there is no single `right answer'. Some people may have longstanding associations with a given place `soaked in memories and meanings'~\citep[][p. 1]{lynch1960}, others may have never visited. The subjective answers that people give for certain built and natural environmental characteristics (e.g., walkability or greenness) might differ considerably from objectively measured indices~\citep{kothencz2017urban,lotfi2009analyzing}.
Although subjective perception varies between people, that does not make perception less important; on the contrary, `a person's quality of life is dependent on the exogenous (objective) facts of his or her life and the endogenous (subjective) perception he or she has of these factors and of himself or herself'~\citep[][p. 136]{dissart2000}. Capturing data on subjective perception gives researchers the opportunity to study the correlation or differences between such exogenous facts and endogenous perceptions; for example, urban development researchers could investigate questions like: do crime statistics in urban areas correspond with perceived safety in those same places? Or, for sustainable transport researchers: is there actually more walking and/or cycling activity measured on streets that people perceive to be more walkable and/or bikeable? The answers to these questions could be used to guide future planning and development policy, leading to substantial improvement in people's perception of their built environment and subsequently their quality of life.

Nonetheless, most urban design, transportation planning and environmental epidemiology researchers rely on objectively measured streetscape indicators because it is easier to collect measurable and quantifiable physical attributes, such as for indicators of green spaces~\citep{kothencz2017urban,labib2020spatial,liu2024clarity}.
However, people's subjective feelings about places do not necessarily correlate with such objectively measured spatial indicators~\citep{mccrea2006strength}. To capture those sentiments as usable data, researchers typically use resource-intensive methods such as field interviews or questionnaires~\citep{lynch1960,lotfi2009analyzing}.
Therefore, we argue that developing a readily deployable method to capture human perceptions of urban environments would provide a much needed tool for answering questions like those posed above. Taken further, these answers can help with efforts to achieve the United Nations Sustainable Development Goals\footnote{\url{www.un.org/sustainabledevelopment/sustainable-development-goals/}} such as Goal 11, `Sustainable Cities and Communities'.

Over the past decade, various companies, organisations and people have been collecting Street View Imagery (SVI) since Google began in 2007~\citep{vincent2007} and the resulting datasets have been widely studied~\citep{biljecki2021svi}.
However, even after all this time, we find challenges remain in the use of SVI for research purposes. In particular, we focus on two major classes of SVI currently accessible to researchers: (i) commercial SVI such as offered by Google, Bing, Baidu or Tencent and (ii) Volunteered Street View Imagery (VSVI) from open platforms such as Mapillary and KartaView (formerly OpenStreetCam), which collect geo-tagged volunteer-submitted photographs and make them available from a central repository.

Unlike commercial SVI platforms like Google, which publish their content under licenses that prohibit certain research methods, VSVI platforms such as Mapillary permit usage of their imagery under the Creative Commons Attribution-ShareAlike 4.0 International\footnote{\url{creativecommons.org/licenses/by-sa/4.0/deed.en}} license, making it easy to adapt into an open-source software system for collecting human perception responses. A common complaint is that the quality of VSVI can vary considerably from very poor to excellent~\citep{ma2019mapillary}, but this problem can be ameliorated by processing and filtering~\citep{zheng2023}. However, no studies have yet developed an open-source toolkit that downloads, filters and processes VSVI and configures a mobile (web) app-based survey for human perception research using it.

Considering this gap, we aim to develop a human perception survey toolkit based on the \emph{findable, accessible, interoperable and reusable} (FAIR) principles~\citep{Wilkinson2016,barton2022making} that takes a \emph{citizen science} approach, which means it enables and encourages non-scientists to collaborate with scientists and contribute towards the advancement of scientific research~\citep{fraisl2022citizen,haklay2015citizen}.
This paper describes the detailed process of creating our open-source VSVI-integrated software aligned with the FAIR principles for Research Software~\citep{hong2022fair}, and provides an example of using this toolkit for perception data collection as a form of participatory sensing~\citep{haklay2015citizen}.
The survey is a mobile-friendly web application that allows participants to rate SVI based on what they see in the image. Participants may be anyone with access to a computer or a smartphone connected to the Internet. Our software presents a lightweight `game-like' user interface~\citep{bakhanova2020targeting} that requires no more and no less than one swipe or button press per image.
The source code for our survey frontend, server backend, and the VSVI filtering and pre-processing scripts, are made available under the GNU General Public License v3.0 and may be found at our Spatial Data Science and GEO AI Lab web site (see the {Software \& data availability} section). Together these constitute a pipeline, suitable for civic/community science~\citep{haklay2015citizen}, that can be assembled by any person with basic knowledge of deploying open source software on a hosted server environment supporting ReactJS, such as virtual servers readily available for low cost from numerous providers.

\section{Background}
Since the launch of Google Street View in 2007~\citep{vincent2007}, the resulting datasets of SVI have been effectively used in numerous studies over a wide variety of domains from urban planning to public health~\citep{biljecki2021svi}. Some of the earliest research validating the usefulness of virtual streetscape audits in place of physical audits~\citep{badland2010}. Looking back at a collection of more recent work over the past decade, \citet[][pp. 16--17]{dai2024street} found that SVI `performs exceptionally well in capturing environmental variables' and `holds immense potential to facilitate environmental health-related studies in the big data era'. Mapillary is a major source of VSVI and includes high-quality panoramic images. With those there is the possibility to crop subimages based on street morphology~\citep{beaucamp2022whole}. However, Mapillary also contains many non-panoramic images and the quality of those can be highly variable~\citep{hou2022comprehensive,ma2019mapillary}. Recent work by \citet{zheng2023} has highlighted ways to improve the usability of VSVI by undergoing a process of filtering, while \citet{biljecki2023sensitivity} found that the use of non-panoramic images only slightly detracted from their results for greenness and sky view visibility indices in the cases they studied compared to the same tests run on panoramic images. \citet{hou2022comprehensive} developed a framework for defining SVI and a consistent method to evaluate its quality. \citet{ding2021towards} used Mapillary image sequences to find bikeway networks in Malm\"o using sign detection machine learning techniques and \citet{sanchez2024accessing} has presented a toolkit using Mapillary images for measuring greenness visibility at eye level.

Traditional approaches to capturing human perception of space include conducting direct observations such as field surveys and resident interviews~\citep{lynch1960}. For instance, `walk-along interviews' may be used to capture the experience of space on the spot while interviewing individuals~\citep{carpiano2009,rzotkiewicz2018systematic}. Field audits and environmental scans are two more tools for evaluating streets. For example, \citet{van2012linking} conducted a field audit to assess the visible greenness level on the streets of two urban neighbourhoods in Gent (Belgium). \citet{harden2024utility} compared the effectiveness of SVI-based `virtual audits' against more traditional in-person environmental scans for assessing `runnability', work that echoes one of the earliest pieces of SVI-based research: comparing virtual against physical streetscape audits for walking and cycling suitability~\citep{badland2010}. Although these approaches provide a more localized assessment of human perception and lived experience of spaces, they are limited due to very high resource and time requirements placed on individual researchers or auditors, even when the audits are conducted virtually or through photographs~\citep{dai2024street}. This human-intensive method does not scale to a large number of places. Considering such limitations, researchers have innovated new ways to survey human perception using massive imagery sets~\citep{dai2024street} and a `crowdsourcing' approach~\citep{bubalo2019crowdsourcing}. We take a closer look below at several recent works that are representative of the main methods used in such research.
Several studies have used a method of scoring imagery based on \emph{relative (pairwise) comparisons}, such as \citet{ye2019visual}, \citet{larkin2022measuring}, Streetscore~\citep{naik2014streetscore} and most notably Place Pulse~\citep{dubey2016deep,salesses2012place}. The last was an extensive and wide-ranging effort to gather a crowdsourced SVI perception comparison dataset; in its second edition, the researchers collected 1.5 million comparisons for 110,998 Google SVI photos. As each participant worked through the web-based survey, they were shown two randomly-selected images at a time, side-by-side, and asked to choose which one of the two was `better' (or equal) according to some selected criterion or `perceptual attribute' such as \emph{Safety} or \emph{Beauty}. These relative comparisons were then translated into absolute scores and an overall image ranking per criterion. The crucial problem of Place Pulse is that to function correctly such comparison-to-score translation algorithms require an order of magnitude more comparisons than the survey was able to collect from human participants. \citet[][p. 6]{dubey2016deep} states that they required `24 to 36 comparisons per image' but could only collect on average 3.35 comparisons per image. Therefore, they developed a customized machine learning algorithm to synthesize additional comparisons based on the collected ones. All this entails a substantial amount of work just to arrive at an absolute score ranking of the SVI in their dataset. Their root justification for using this relative comparison method comes from psychological research that studied humans performing tasks of making absolute identifications vs relative comparisons of simple stimuli such as `sound tones' or `line lengths'~\citep{stewart2005absolute}. However, a significant limitation of that earlier psychological research is that it did not consider the evaluation of complex imagery such as SVI nor did it ask participants to rate imagery based on higher-level conceptual criteria such as `safety' or `walkability'.
In contrast, \citet{twedt2016} used the direct \emph{absolute scoring} approach, with a survey website where each image was shown one at a time and had to be rated on a scale of 0 to 100 using a slider adjusted with the mouse. Only 40 images were rated overall. There were approximately 300 participants, each was paid a small amount for their time via Amazon Mechanical Turk, and shown a preview of all the images before being asked to rate them individually. \citet{pearson2024} operated similarly but scaled up to 10,727 paid participants, each of whom rated 33 images. They reported that the actual average completion time for the survey was 104 minutes, which was considerably higher than the 8 minute completion time they had expected from internal testing. %
\citet{kruse2021places} paid 210 Mechanical Turk workers to consider `playability' on a five-point absolute scale and rated 3,011 images from 3 U.S. cities.

\citet{yao2019human} created a human/machine feedback-loop for speeding up manual absolute scoring work while simultaneously improving the accuracy of modelled perception ratings, on the same perceptual attributes as Place Pulse (e.g. Beauty, Safety, etc).
This approach sped up the manual rating process to approximately 1,000 images per hour (3.6s per image) using a recommendation algorithm based on a machine learning model.
However, there are several problems with both the method and the design of the software.
Firstly, one of the biggest barriers to reuse of this survey method is that it is a software application that must be downloaded and installed locally on each participant's computer.
This was not much of an obstacle for the authors because all of their participants were invited students or university staff, who were presumably comfortable with installing software from a fellow university member, but it is a major problem when trying to attract participation from the wider public.
Secondly, the software's user interface is not user-friendly because of the amount of mouse-movement required for each image: the participant adjusts a slider to a value between 1 and 100 and then they must activate a submit button (although at least some of this interaction is also possible via hot-key bindings). Reducing this user-interaction effort appears to be one of the major underlying motivations of this work.
However, this leads to the third problem: to speed up the rating process, the recommendation algorithm chooses a rating that is ready to be submitted by default, but in doing so it rewards participants who simply agree with the recommendation. There is no way to distinguish true human ratings from computer-generated ratings for which the human participant did not have strong enough feelings to justify the effort of moving the mouse to change the slider away from the default setting. Therefore the algorithm risks tainting its own training data too much.
Fourthly, the authors noted a significant problem, which we see as being related to (but not quite the same as) the third: the given recommendation may have the unintended consequence of influencing the participant even when they do choose to adjust the slider. For example, suppose a participant would have rated an image with a score of 20 in Beauty, but then they saw the machine recommended a score of 40, so they decided to compromise on a score of 30. There is still an element of human input in this case, but it is biased by the recommendation algorithm.   %

The user interface presented by our survey app utilizes a game-like `design pattern'~\citep{morschheuser2017gamified} with large icons that can be swiped, and congratulatory messages at various stages of progress, but does not proceed any further down that gamification path. It is not as elaborate as MapSwipe~\citep{ullah2023assessing} or Tomnod~\citep{baruch2016motivations}, both of which involve crowdsourcing spatial data that is rendered on a map, but rather more akin to Galaxy Zoo~\citep{masters2019twelve}, in which images are categorized by clicking a button (in that case, by galaxy type) in a slick and customized (web) app.

All of these aforementioned SVI-based works relied on either author-supplied photographs or commercial SVI, usually Google, but some used Tencent or Baidu. We instead set out to build reusable, open and FAIR software for assembling and operating street view perception surveys based on an open provider of VSVI, Mapillary.

\section{Materials and methods}

\subsection{Mapillary street view imagery}

\begin{figure}[tb]
    \centering
    \includegraphics[width=0.45\textwidth]{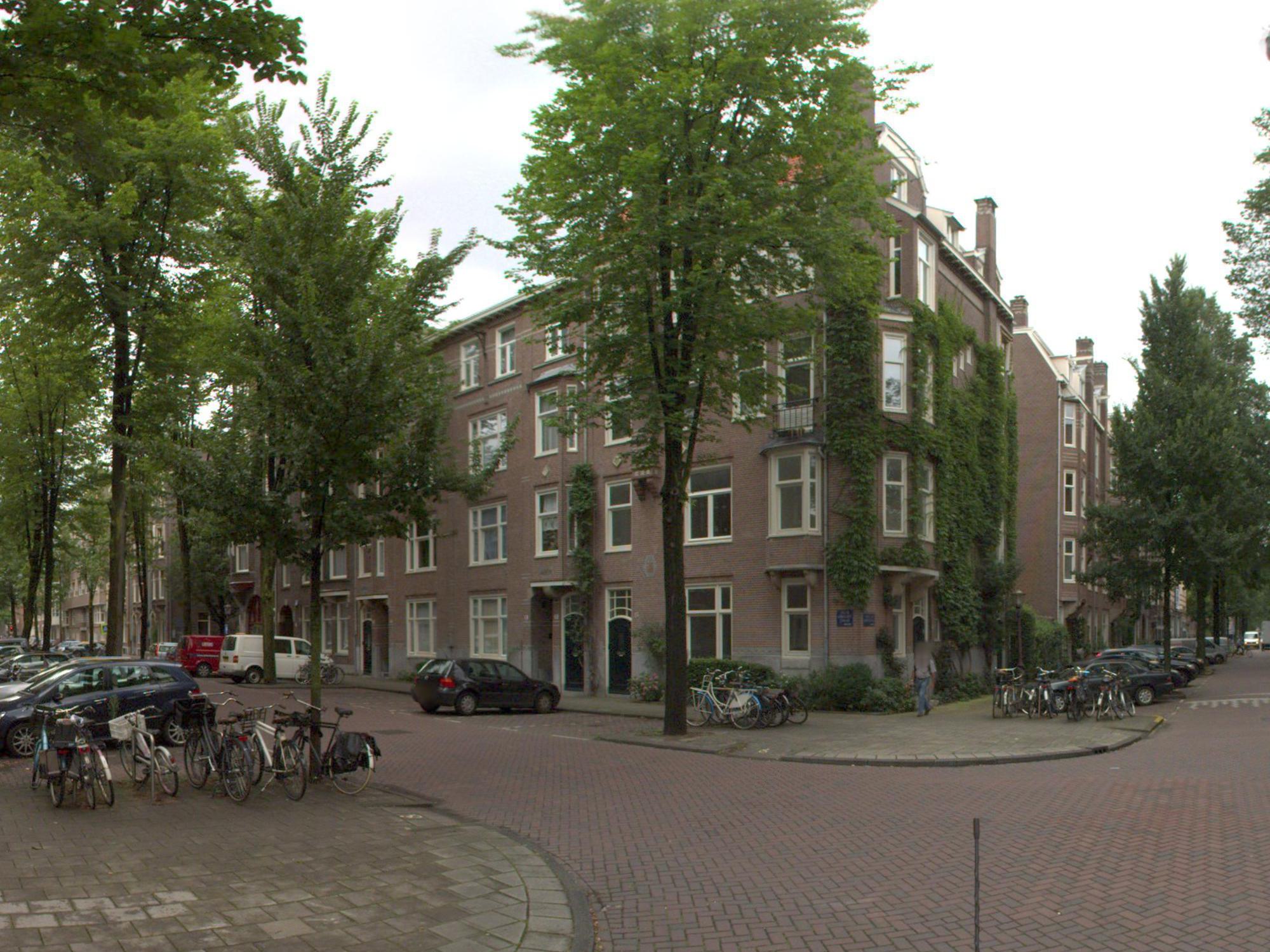}
    \caption{Sample image from Amsterdam (source: Mapillary)}\label{fig:sampleimageams}
\end{figure}
Our method uses imagery from the Mapillary platform, which offers free-to-use SVI and VSVI in many cities around the world under the terms of a Creative Commons license. Some of this imagery was collected professionally with high-quality 360-degree panoramic cameras, especially in cities like Amsterdam (see Figure~\ref{fig:sampleimageams} for a sample of a cropped subimage of a panoramic photograph). Most of the available imagery is VSVI, of varying quality, but recent work by \citet{zheng2023} shows that such VSVI can be usefully filtered and employed in research as an alternative to higher-cost or non-free options, while \citet{biljecki2023sensitivity} found that working with non-panoramic imagery could bring them reasonably close to deriving the same results as working with panoramic imagery in the cases they studied. All of Mapillary's available imagery is processed with face- and license plate-blurring software according to their privacy policy\footnote{\url{www.mapillary.com/privacy}}. However, whichever SVI provider a researcher chooses to use, they should ensure that the images are appropriate, ethically sound and privacy protected, as they would with any other data source.

The motivation to use Mapillary VSVI comes from the FAIR data movement~\citep{Wilkinson2016}. Guidance for the final FAIR principle, \emph{reusability}, emphasizes that data must be not only technically interoperable but also legally interoperable\footnote{\url{www.go-fair.org/fair-principles/}, section R1.1}. However, the standard terms of service\footnote{\url{cloud.google.com/maps-platform/terms/}} for Google SVI stresses that images are non-free and explicitly forbid the kind of usage that we need for research. In particular, in section 3.2.3 of the terms of service, they prohibit \emph{scraping}, \emph{pre-fetching}, \emph{bulk downloading}, \emph{storing}, and \emph{resharing} Google SVI, all of which are necessary components of our survey method. Bing, Baidu and Tencent Maps also list several similar restrictions in their terms of service, and in the case of the latter two, those platforms are mainly focused on providing imagery from China rather than worldwide.

While Google imagery is high-quality and widely available, we intend to keep within the open and FAIR principles. In particular, we are concerned that some people wishing to use our toolkit may not be able to obtain the authorization to use Google imagery, or to produce freely-reusable datasets from it, such as those who work in countries where there is legal uncertainty about these ways of using proprietary data~\citep{helbich2024proprietary}. We simply wish to show that there are alternatives that are free in the sense of being accessible and available (`\emph{libre}') for research usage without significant restrictions on reusability. The fact that Mapillary is free in the sense of monetary cost (`\emph{gratis}') is also helpful. In contrast to Google, the Mapillary terms of service\footnote{\url{www.mapillary.com/terms}} allows usage of VSVI under a Creative Commons license that only require proper attribution of imagery that is downloaded and reshared or integrated into applications. We also note that there are some countries for which Google SVI is sparse or barely available at all, while Mapillary has substantially more VSVI in those same countries that can be reasonably filtered and processed; e.g., as of September 2024 this is the case in Zambia, Morocco and Nicaragua, among others.

\subsection{Imagery downloading and processing}

Our Python script, \filename{mapillary\_jpg\_download.py} gathers imagery from Mapillary via their API (see Appendix~\ref{sec:mapillarydl} for more details). The script only requires the acquisition of a free API key from the Mapillary developer portal, and to be given a geographic bounding rectangle for some region of interest. It is also possible to obtain imagery from other sources, and inject it into the process at this point, should Mapillary be unsuitable or undesired for any reason, however the focus of this work is on Mapillary-provided imagery.

Once the tile data files and photographic imagery are downloaded, we must process a very large number of JPEG files, some of which are panoramic photographs projected into a wide image format, and others which are simply plain photographs. The two main tasks we need to accomplish are: (1) finding `sensible subimages' in the panoramic imagery to crop out and save separately, and (2) weeding out any
images that are too dark, too blurry or defective in some other way (such as only showing a wall or an undifferentiated block of greenery). These are both somewhat vague needs but thanks to recent advances in machine learning they can both be addressed using off-the-shelf software.
In the case of panoramic photographs, our toolkit automatically crops subimages of a 4:3 ratio to show users images that fit within the app's available view-port for imagery, rather than distorted and overly large raw panoramic images.

\begin{figure*}[tb!]
    \centering
    \includegraphics[width=\textwidth]{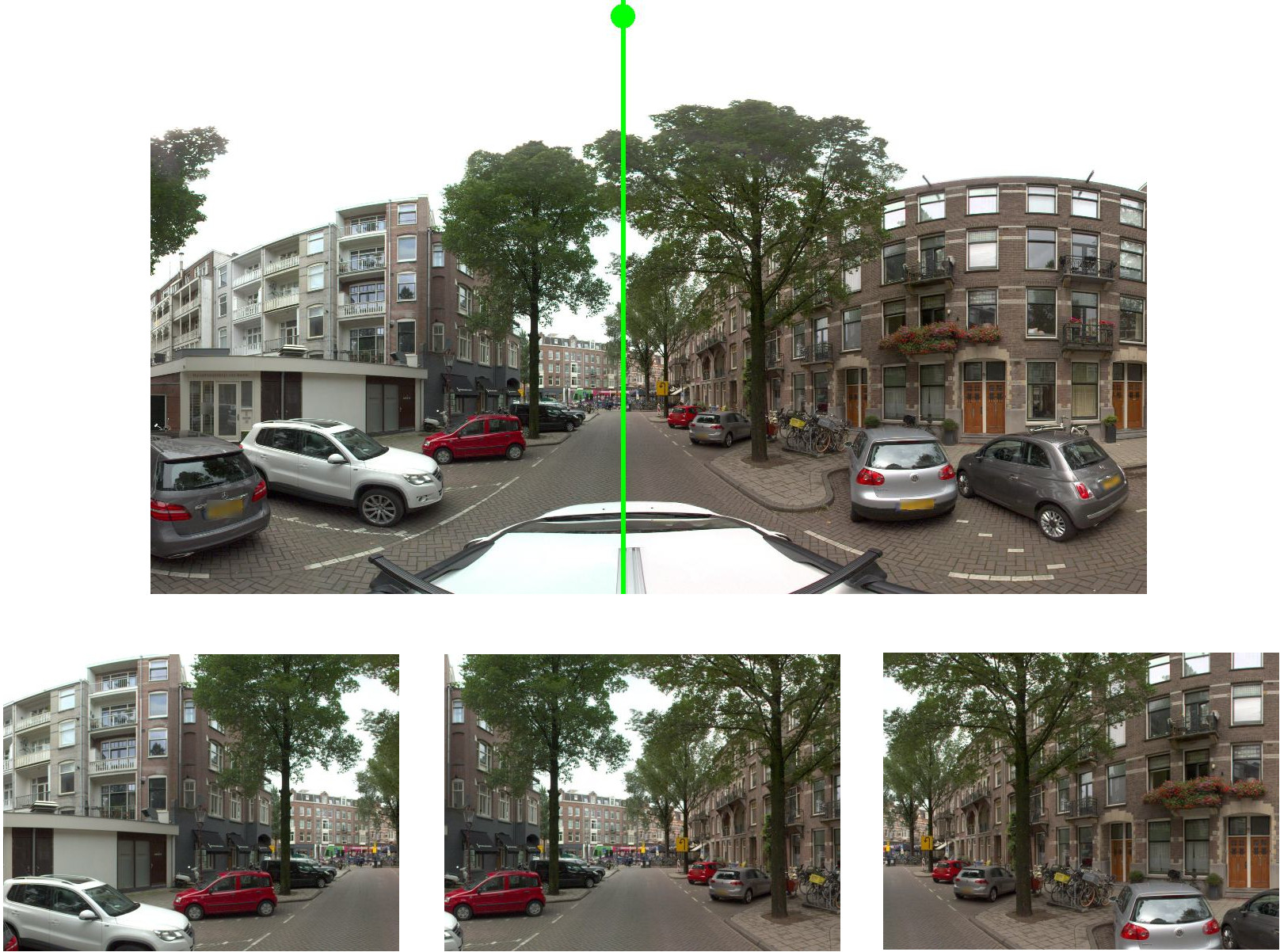}
        \caption{For each road center (indicated by the green line that is also marked with a circle) found in a panoramic image (top), we take three crops (bottom, from left to right): one slightly left of center, one directly facing the center, and one slightly right of center.}
    \label{fig:samplecrops1}
\end{figure*}

Our definition of `sensible subimages' was refined through trial and experiment, and we settled on the following: find the centers of roads in each of the panoramic images, and then for each road center crop a series of 4:3 images slightly to the left of center, on the center itself, and slightly to the right of center, as shown in Figure~\ref{fig:samplecrops1}. This captures a wide variety of ways of looking at streets, including many views with a great deal of built environment and greenery, as well as views straight down the center of roads. \citet{kim2021decoding} found that these choices of viewing direction, when sampling from panoramic SVI, could lead to large differences in terms of streetscape measures. Therefore, it is important for toolkit users to carefully consider the location of cropped subimages within panoramic SVI and how that might affect the results of their survey.

\paragraph{Finding road center-lines}
The panoramic imagery from Mapillary is generally normalized so that the leftmost edge of the image is where the compass would point north from the location where the SVI was taken. In theory, it should be possible to reconstruct the position of streets within panoramic imagery based on street map vector data (such as OpenStreetMap). However, in practice, we found some discrepancies when examining SVI samples: the due north direction was not always the leftmost edge. In addition, the arrangement of streets in reality can be considerably more complex than shown in the simplified model of a vector-based street map. By finding road center lines using a computer vision algorithm, our software is robust in the case of imagery where compass angle is wrong or simply not known.
We find the center of roads in imagery by first labeling all of the pixels as either `road' or `non-road' using semantic segmentation~\citep{thisanke2023}, and then seeking the `peaks' of the road pixel distribution horizontally across the image, with some additional code to handle common distortions and edge cases. See Appendix~\ref{sec:roadcenters} for more details.

\subsection{Filtering by image quality}\label{sec:filtering}

Most of the volunteered photographs that we encountered on Mapillary, while looking at various cities around the world, are non-panoramic and can range from very good quality to quite poor. For example, some unusable images we found barely showed anything at all but a single wall, a blur of greenery or a close-in view of a parked car.  Inspired by the filtering ideas of \citet{zheng2023}, we calculate contrast and `tone-mapping score'~\citep{stefanescu2021imagequality} using the Python Scikit-image library~\citep{van2014scikit}, and apply a certain threshold on a formula composing both values (see Appendix~\ref{sec:qualitythreshold}), to select acceptable images. We also applied the road center-line finding algorithm described above, but in this case, only to determine if there is a road or not within the image; through trial and error, we found this simple `road check' to be a good heuristic that met our need to capture an open and clear view of a street and the surroundings that a person would see if they were standing there. Images that pass these tests are then automatically cropped by our toolkit to a 4:3 ratio so they fit into the image view-port of our web app. However, all of these the filtering criteria can be adjusted, or modified entirely, to suit any particular research need. For example, if a researcher would like to include `non-road' images in their dataset, then that is possible by turning off the road check.

It is possible for so-called `bad' images to slip through, ones that have poor photo quality or some form of obstruction that renders them ill-suited for evaluation; however after experimentation we found that such instances are rare, and they can be reported when found by a participant. To some extent there is also a subjective feeling of `badness', which cannot be controlled. For example, some people prefer images with blue skies and do not like rating SVI taken on cloudy days, whereas other people do not mind such images. Since the survey is purely voluntary, we permit people to skip images that they feel they cannot rate, for any reason, as to not discourage participation.

\begin{figure}[tb!]
    \centering
    \includegraphics[width=\textwidth]{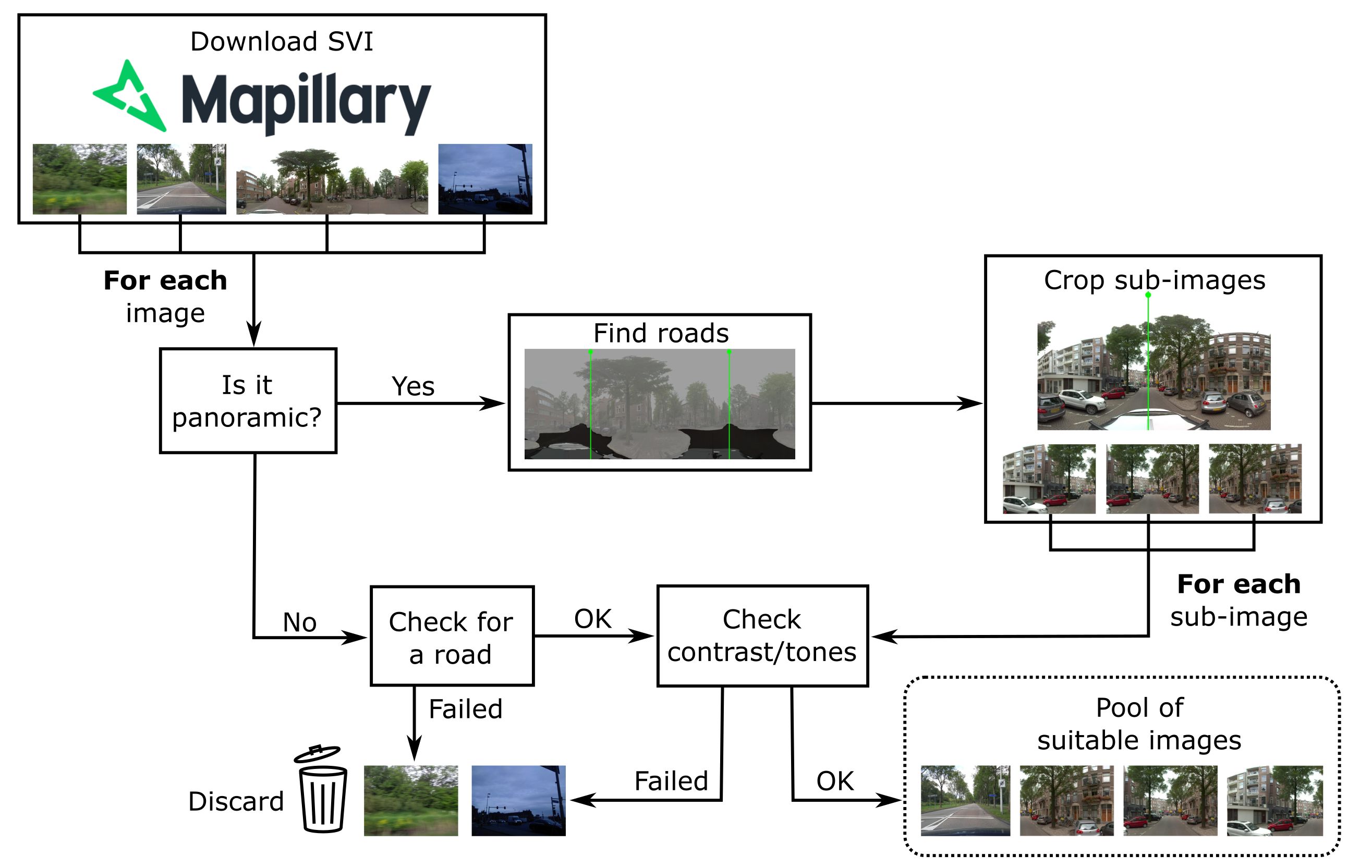}
    \caption{Flow diagram of the preparation and processing of SVI}
    \label{fig:prepdiag}
\end{figure}

An overview of all the steps of downloading and processing of Mapillary VSVI are shown in the diagram within Figure~\ref{fig:prepdiag}. After these steps, the imagery is ready to be used in the perception-gathering survey, or it may be taken and used for other purposes in compliance with Mapillary terms of service.

\subsection{Survey frontend}

\begin{figure*}[tb]
    \centering
    \includegraphics[width=0.4\textwidth]{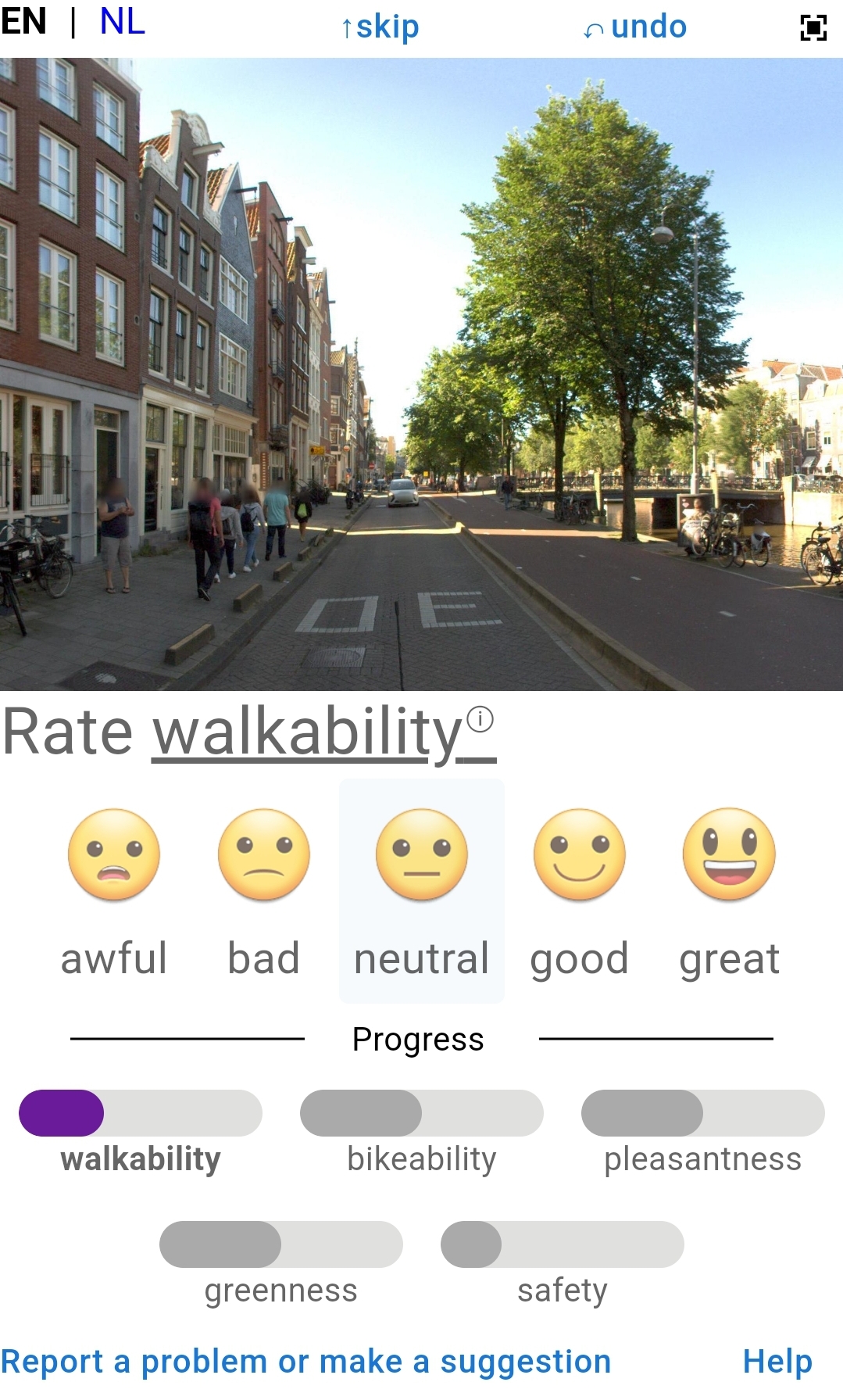}\quad
    \includegraphics[width=0.4\textwidth]{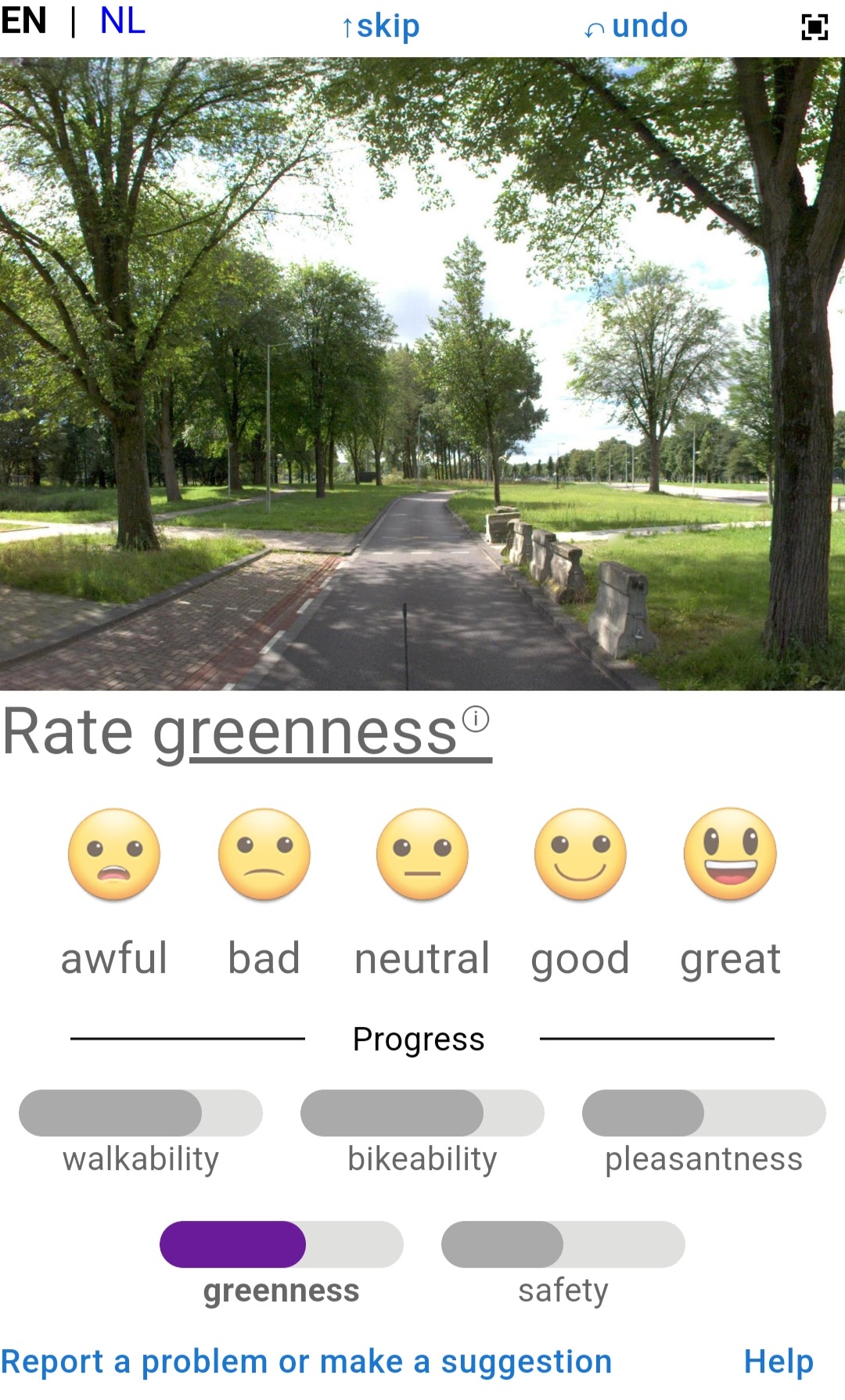}
    \caption{Two examples of mobile screenshots}\label{fig:mobilescreenshots}
\end{figure*}
The perception-gathering survey is written as a mobile web app in the ReactJS framework. First of all, before the main perception-gathering survey takes place, there is a short socio-demographic survey requesting a few personal details: age, education level, gender, approximate income, home postal code, country of residence and consent for data collection (see Section~\ref{sec:demographics} for more details). Only age and consent are required fields. We comply with the EU General Data Protection Regulations\footnote{\url{commission.europa.eu/law/law-topic/data-protection_en}} and require participants' explicit consent, which may be withdrawn at any time. Furthermore, we do not use data from participants who are under 18.

The main perception-gathering survey then proceeds: it shows participants one image at a time along with one of five possible categories: walkability, bikeability, pleasantness, greenness or safety (see Appendix~\ref{sec:categories} for more details). Participants may rate each image according to the given category by pressing one of five rating buttons, or swiping the image towards one of the rating buttons. The buttons are arranged along a Likert-type scale: awful, bad, neutral, good, and great (internally numbered from 1 to 5). Participants rate five images according to a given category and then the app chooses a new category at random, until 20 images have been rated in each category. There are also options to skip the current image, undo the last rating (see Appendix~\ref{sec:undoprotocol} for more undo details), go full screen, change language, get more help, and report an issue. The mobile web app (Figure~\ref{fig:mobilescreenshots}) is designed to fit seamlessly on a mobile device so that users can pull up their smartphone to quickly do some ratings whenever they desire. Our decision to collect absolute scores (instead of pairwise comparisons) makes it possible for us to present a clean and simple user interface to participants, with an intuitive swipe-to-rate input mechanism.

The frontend runs entirely on the user's browser and communicates with the backend server via a public API (see Appendix~\ref{sec:publicapi} for more API details). Image URLs are fetched from the backend, and then downloaded from the image-hosting server as needed. Ratings and undo commands are sent asynchronously to the backend while the interface updates. Text in the user interface is managed by the \verb+react-intl-universal+ module so that we (or other contributors) can easily add new language translations and locales when preparing to deploy the survey in different countries.

The user interface was pilot-tested with Utrecht University students and staff. We made several changes as a result, including: the positioning of the SVI, the text of the category descriptions and the wording of the socio-demographic survey items. We also decided to limit participants to 20 ratings per category (100 ratings total) and show progress bars in each category to give people a sense that they were working towards a definite ending.

\subsection{Survey backend}

\begin{figure}[tb!]
    \centering
    \includegraphics[width=0.75\textwidth]{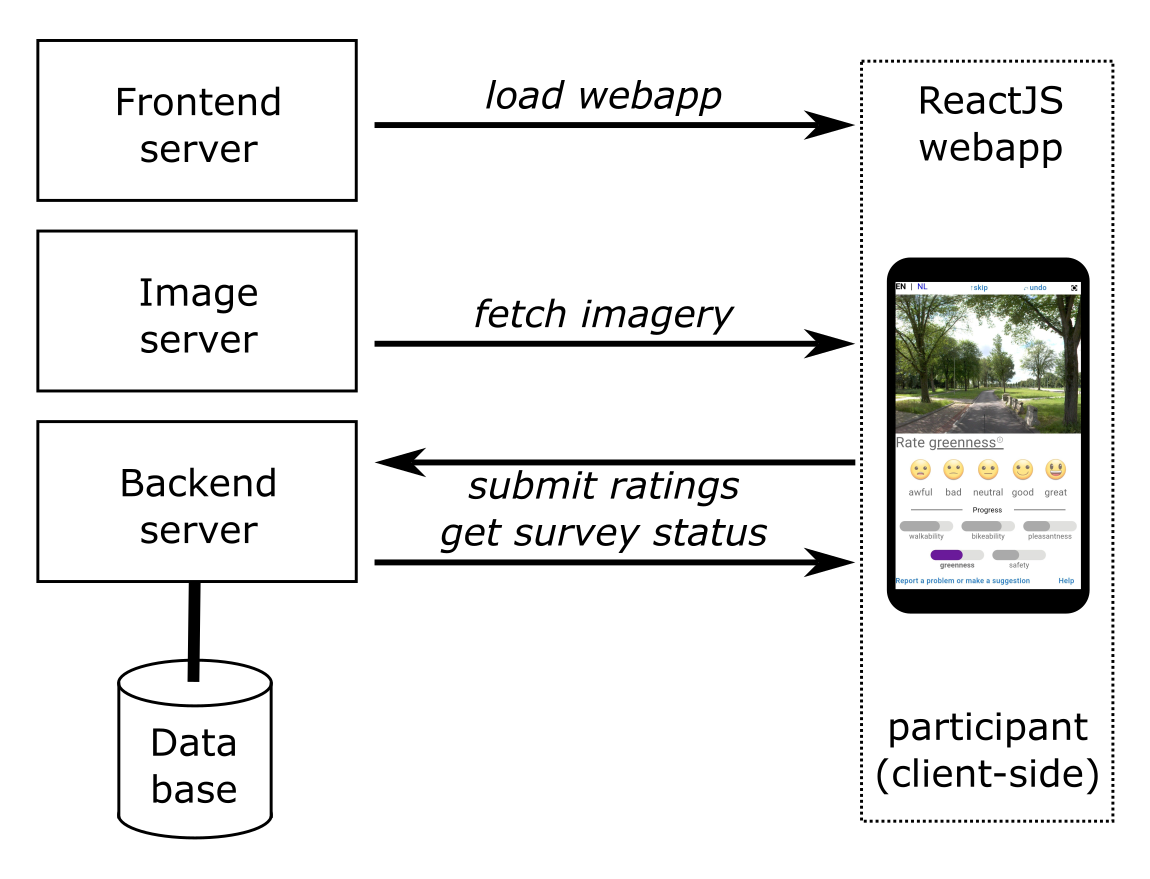}
    \caption{A schematic showing server-side (left) / client-side (right) and their interactions.}\label{fig:sysdesign}
\end{figure}

The backend is an independent ExpressJS-based server with a well-defined public API for tasks such as starting new sessions and submitting surveys and ratings. The backend is a separate module from the frontend; it is possible that a different frontend (or even raw requests) could interface with the backend, if so desired. The public API (see Section~\ref{sec:publicapi}) is treated as a potential entry point for input from any possible source, including malicious ones. Data are stored securely on a private server using PostgreSQL and PostGIS for geographic data processing, where they are kept until they have been suitably anonymized or stripped of personally-identifiable information in a manner that is appropriate for publication. The overall arrangement of frontend, backend and client (in this case, depicted as a smartphone user) is shown in Figure~\ref{fig:sysdesign}.

\subsection{Computational Resources}
The survey server that we used for our case study is a Linux virtual machine hosted on an Intel Xeon\texttrademark\ 5120 CPU with 4 cores running at 2.2GHz, having 16GB RAM available and approximately 100GB of disk space used for data and imagery. Reasonably comparable virtual servers can be hired (as of this writing) for around \euro 10-20 per month. We had no more than 13 people simultaneously participating in our survey and our server had more than sufficient resources to support that load. In terms of hardware and software flexibility, it should be possible to run our open source software on any operating system that supports ReactJS v18.4 and the PostgreSQL v13 database, however we have only tested (and therefore recommend) the following: Linux (CentOS 7) with Apache2 v2.4.6 (configured as a proxy), PostgreSQL v13.14 with PostGIS v3.3.3, and Node v18.4.2 with React v18.2.0 and Express v4.18.2. Newer versions of dependencies should work fine, especially regarding Apache2 and PostgreSQL (which on CentOS are substantially older but stable versions maintained by the vendor). Code changes may be necessary for updated Node and React libraries and that will be managed on our GitHub repository. Preprocessing operations were handled by Python 3.6 scripts under Linux but should work fine with newer versions. We recommend having at least approximately 2 TB of disk space available for comfortably managing to download and filter VSVI on a region comparable to the one we describe in Section~\ref{sec:surveysetup}; this larger set of raw VSVI can later be deleted to save space if desired.

\section{Results}

\subsection{Survey set-up for the Amsterdam case study}\label{sec:surveysetup}

The primary purpose of this paper is to introduce our SVI survey toolkit, however we conducted a case study as a demonstration that also collected some useful data. We retrieved imagery from points found within the bounding box described by longitudes 4.7149 and 5.1220, and latitudes 52.2818 and 52.4284 (WGS 84). This encompasses the city of Amsterdam and some outlying areas. The total number of images found (panoramic and otherwise) in this bounding box was well over 700,000. The panoramic images available from Mapillary for Amsterdam are almost entirely high-resolution and professional-quality, taken from a 360-degree panoramic camera mounted on a vehicle or backpack. Using our method, each panoramic image could potentially be used to derive up to nine high-quality subimages. With so many possibilities, we heavily filtered the amount of possible SVI first by applying the pipeline as described in Section~\ref{sec:filtering}, and then selecting only images from locations closest to a fixed geographic grid of points covering the whole region but spaced approximately 20m apart from each other. This still left us with too many images, so we then down-sampled from these eligible images randomly until the number of images for the survey was a bit under 20,000, which we felt was a sufficient number for our initial run, and well-distributed across the region of interest. This is effectively increasing the sparsity of sampled points, which \citet{kim2021decoding} found to lead to higher variability in streetscape measures; however, it was not practical to conduct our case study with significantly more images because we were not realistically expecting many tens of thousands or more responses.

\subsection{Collected perception}

\newcommand\II{\hspace{1em}}
\newcommand\SDC[1]{\textbf{#1}}

\begin{table}[tbp]
\begin{tabular}{lc}
description & percentage of participants (actual number) \\ \hline
\SDC{total participants} & 100\% (\Stat{totalparticipants}) \\
\SDC{gender} \\
\II woman & \Stat{genderwoman} \\
\II non-binary & \Stat{gendernonbinary} \\
\II man & \Stat{genderman} \\
\II \em other or not specified & \Stat{genderother} \\
\SDC{education level} \\
\II postgraduate & \Stat{edupostgraduate} \\
\II tertiary & \Stat{edutertiary} \\
\II secondary & \Stat{edusecondary} \\
\II primary & \Stat{eduprimary} \\
\II \em not specified & \Stat{eduother} \\
\SDC{age} (average: \Stat{ageavg}, standard deviation: \Stat{agestddev}) \\
\II 18--27 & \Stat{age1827} \\
\II 28--37 & \Stat{age2837} \\
\II 38--47 & \Stat{age3847} \\
\II 48--57 & \Stat{age4857} \\
\II 58--67 & \Stat{age5867} \\
\II 68--77 & \Stat{age6877} \\
\SDC{country of residence} \\
\II{The Netherlands} & \Stat{nlparticipants} \\
\II{Another EU country} & \Stat{euparticipants} \\
\II{A non-EU country} & \Stat{noneuparticipants} \\
\II\em not specified & \Stat{otherparticipants} \\
\SDC{estimated monthly gross income (mgi)} \\
\II \euro 0--1,999 & \Stat{income00001999} \\
\II \euro 2,000--3,999 & \Stat{income20003999} \\
\II \euro 4,000+ & \Stat{income400015000} \\
\II\em not specified & \Stat{incomeother} \\
\end{tabular}
\caption{Socio-demographic data summary}\label{tab:sociodem}
\end{table}

\newcommand\code[1]{\texttt{#1}}
\begin{table}[tbp]
\scriptsize
\begin{tabular}{cccccccccccc}
\code{id} & \code{timestamp} & \code{sess} & \code{image} & \code{cat} & \code{score} & \code{postcode} & \code{country} & \code{age} & \code{mgi} & \code{education} &   \code{gender} \\
\hline
8710 & 2023-06-27 16:21:03 &        145 &    77114 &           3 &      4 & 3\verb+--- --+& Netherlands &  26 & 2300 & Postgraduate & man \\
8719 & 2023-06-27 16:21:23 &        151 &    85673 &           2 &      3 & 3\verb+---+& Netherlands &  38 & 4000 & Postgraduate & woman \\
8976 & 2023-06-27 16:43:48 &        151 &    71732 &           1 &      4 & 3\verb+---+& Netherlands &  38 & 4000 & Postgraduate & woman \\
13378 & 2023-07-12 15:05:25  &        211 &    42535 &           4 &      2 & 1\verb+--- --+& Nederland &  35 & 3500  & Tertiary  & man \\
13460 & 2023-07-12 22:47:45 &        212 &    48075 &           4 &      2 & 1\verb+--- --+& netyerlands &  58 & 5000 & Postgraduate & woman \\
19100 & 2023-09-12 10:33:38 &        301 &   190126 &           4 &      4 & 1\verb+---+& Belgium &  18 & 200 & Secondary & non-binary \\
\ldots
\end{tabular}
\raggedright
\normalsize
where\\
\begin{tabular}{r@{: }l}
\code{id} & A unique identifier assigned to every rating submitted by any participant. \\
\code{timestamp} & The date and time of the submission (recorded to the microsecond, not shown here) \\
\code{sess} & The unique identifier assigned to every participant when they start a rating session. \\
\code{image} & The unique identifer (previously configured in our database) of the image that was rated. \\
\code{cat} & The rating category: 1=Walkability, 2=Bikeability, 3=Pleasantness, 4=Greenness, 5=Safety. \\
\code{score} & The submitted rating: 1=awful, 2=bad, 3=neutral, 4=good, 5=great. \\
\code{postcode} & Home postal code of the participant (redacted here). Free-form entry in the survey. \\
\code{country} & Country of the participant. Free-form entry in the survey, hence we must deal with variation. \\
\code{age} & Age of the participant. Aside from consent, the only required entry in the survey, and numeric. \\
\code{mgi} & Estimated monthly gross income. Optional in our survey, localized to the Netherlands (\euro). \\
\code{education} & One of: Primary, Secondary, Tertiary or Postgraduate. Optional in our survey. \\
\code{gender} & One of: woman, non-binary, man, unspecified or free-form text entry.
\end{tabular}
\caption{Data sample. Note that the underlying data is stored in normalized tables, this is a joined-together view for consideration and analysis. The postal codes have been mostly redacted, but their shape remains (with dashes replacing digits and letters) to show the variability of the underlying survey data. The timestamps and ages have been randomly perturbed as well~\citep{rahman2023efficient}. As shown, the raw data can be messy and require some processing to clean up inconsistencies like those found above in the \code{country} column.}\label{tab:datasample}
\end{table}

From April 2023 to February 2024, we received \Stat{totaldatapoints} ratings across \Stat{totalimages} images from up to \Stat{totalparticipants} participants. We recruited participants using a wide variety of channels including but not limited to: social media, institutional mailing lists, and classrooms. We asked several socio-demographic questions of participants before they began the main part of the survey. The responses to the socio-demographic questions are summarized in Table~\ref{tab:sociodem}.

Of our \Stat{totalparticipants} participants, \Stat{completedsurveys50} submitted at least 50 ratings, and \Stat{completedsurveys100} finished the full 100. The median survey completion time was \Stat{mediandurationminutes} minutes and \Stat{duration30} of participants took 30 minutes or less from start to end. We observed participants swiping or clicking with ease through our smartphone-friendly user interface, and we later confirmed this using timestamps in our database: the median interval between ratings was \Stat{medianintervalseconds} seconds and \Stat{intervalseconds10} of the submitted ratings occurred within 10 seconds of the previous rating. A sample of the collected information is shown in Table~\ref{tab:datasample}.

\subsection{Spatial distribution of perception}

\begin{figure}[tb!]\centering
    \includegraphics[width=0.95\textwidth]{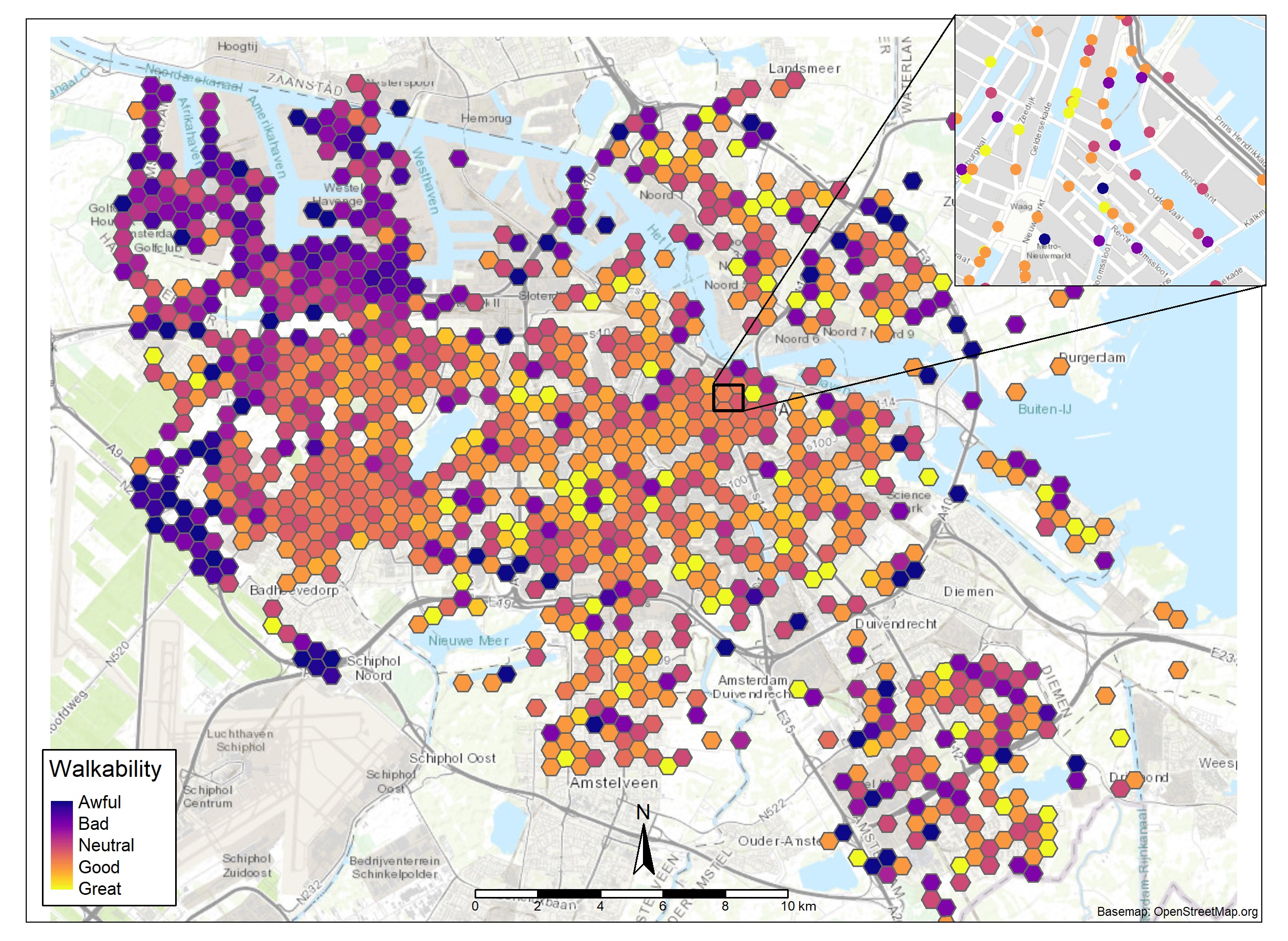}
    \caption{Collected walkability ratings in and around Amsterdam. The inset map shows individual points colored by score within a small region of the city, the larger map shows these points aggregated and averaged into hexagonal bins.}\label{fig:walkabilitymap}
\end{figure}

As an illustrative example, Figure~\ref{fig:walkabilitymap} shows the collected walkability geocoded perception data, averaged into hexagonal bins each measuring approximately 650 m wide west/east and 600 m north/south. These fully-anonymized data points for this map are available on our web site. We find it encouraging that this overview of the data aligns with the common sense intuition that worse walkability perceptions occur more often on the outskirts of the city and better perceptions should be more frequent closer to the center of the city. We note that many of the better ratings that are found in the outskirts tend to be linked to photos of residential neighbourhoods or parks, and the worse ratings with industrial areas or high-speed roads, which are more common outside the center.

\section{Discussion}
\subsection{Main findings}

The goal of this work is to provide an open-source end-to-end pipeline for building and conducting street perception surveys. This enables citizen science-based~\citep{haklay2015citizen} perception research in two ways: firstly, the survey itself is a form of participatory sensing through an easy-to-use smartphone-friendly web app. Secondly, by keeping to the FAIR principles~\citep{Wilkinson2016,barton2022making} for data and software, we promote open science and reproducibility.
Our choice to rely on FAIR data led us to use preparation methods inspired by \citet{zheng2023} for cleaning and selecting VSVI rather than obtaining potentially cleaner SVI from proprietary platforms that have overly restrictive terms of service. 

Our survey app asks the participant to view only a single image at a time, unlike Place Pulse~\citep{dubey2016deep,salesses2012place} or other relative comparison studies~\citep{naik2014streetscore,ye2019visual,larkin2022measuring}, which involve showing the participant at least two images at a time. Like \citet{kruse2021places}, we collect ratings along an absolute five-point scale, whereas much of the challenge in projects such as Place Pulse was in reconstructing a ranking from a large set of relative comparisons; this is a basic philosophical difference in approach.
The advantage of using absolute scores is that the ratings are always linearly ordered, which is helpful when the expected study sizes are on the smaller side, working with only thousands of responses, or even fewer. We also argue that with high-level and high-complexity concepts such as `walkability' and `safety' the use of relative comparisons is not necessarily helpful, since there are so many aspects of the imagery to consider simultaneously; past psychological research to justify the use of relative comparisons~\citep{stewart2005absolute} considered only low-complexity concepts such as `line lengths' or `sound tones'. Instead of having a complex process of translating relative comparisons into absolute scores, we keep things simple in our system by showing only one image and enabling participants to give a quick first-impression one-click/swipe answer from a selection of five possible responses. This user interface helps participants quickly and smoothly look through many more images than interfaces with a rating slider~\citep{twedt2016,larkin2022measuring} and avoids any biasing problems created by feeding back preliminary machine learning model results into the user interface~\citep{yao2019human}, while still allowing participants to able to enjoy rapid progression through the survey.

We were able to easily attract public participants and give them the flexibility to respond at their convenience because our survey is a smartphone-friendly web app rather than a downloaded software application. We did not pay participants~\citep{twedt2016,pearson2024,larkin2022measuring,kruse2021places} but rather relied on the attractiveness of the survey's appearance as an almost `game-like' app.
Our participants took \Stat{medianintervalseconds} sec per image to perform ratings (at the median), which is very close to the 3.6 sec per image timing results reported by \citet{yao2019human}, but without the need for feedback (or interference) from a machine learning model.

\subsection{Limitations}

Mapillary\footnote{\url{mapillary.com/about}, as of September 2024} has more than 2 billion images from 190 countries, and especially good coverage in countries like the Netherlands, including almost full panoramic coverage of Amsterdam-area streets. However, although Mapillary has imagery from every continent, the most abundant SVI coverage comes from Europe and North America~\citep{ma2019mapillary}. However, commercial SVI is no panacea either: \citet{kim2023examination} analyzed Google SVI coverage of walk commute trajectories in small- to medium-sized cities in the United States and found gaps in nearly half of the routes they investigated; they suggest that researchers consider Mapillary to bolster coverage.

Like with any SVI-based measurements, the outputs of the survey created by this system will depend upon imagery selection choices with regard to directionality, spacing and specific viewing position on the street (e.g. sidewalk-view vs road-centered-view) for each image~\citep{kim2021decoding,ki2023bridging}. For the sake of the case study we have made choices specific to the Amsterdam area and the practical considerations we had to make in terms of the level of participation we could reasonably expect. Researchers deploying this toolkit on their own studies will need to consider similar issues specific to their situations.

The five-point scale is simple for participants but as a result sometimes is too coarse-grained to express the true rating that the participant feels. Similarly, descriptions of categories try to give unambiguous criteria for people who need some guidance with their consideration of images, however with some categories there is an unavoidable tension between different interpretations. For example, with `safety' many of the survey testers found themselves thinking of a `road safety' interpretation rather than a `personal safety' interpretation, although we tailored the description to fit the latter. However, we do not wish to dictate the responses from participants, in the end we are trying to collect what people are already thinking rather than trying to teach them something, and it is reasonable to consider road safety as a component of personal safety.

The absolute scoring system avoids the complexity of the relative comparison system used by Place Pulse~\citep{dubey2016deep,salesses2012place} but can result in situations where participants might change their minds about images and wish to go back and redo the ratings. We purposefully limited `undo' functionality to a single previous rating to prevent participants from undoing large numbers of ratings. This protects the integrity of the system against unexpected mass cancellation of ratings. Our system collects `quick impressions' with the idea being that participants spend no more than a few seconds on each image; revision or reconsideration of ratings would work against that goal, even if the participant later thinks differently. Relatedly, the absolute scoring system makes it possible to rank images with trivial effort (because absolute scores form a total order), but that ranking could be subject to significant fluctuations when the number of ratings per image is low. This could happen if new ratings for an image are collected that significantly differ from the existing ratings for that image. Therefore, the potential number of participants, and how many images they might be able to rate, should be considered when choosing the number of images to include.

Our survey relies on Internet access and works best on a modern smartphone although it can be completed on a regular computer with a reasonably up-to-date (within the last half-dozen years) web browser as well. Not everyone has Internet access, although recently published survey work from the Pew Research Center~\citep{poushter2024pew} found that in most countries surveyed approximately 8 out of 10 adults, or more, use the Internet.

The toolkit is currently split into three separate components, each of which can be used independently of the others. Furthermore, as a web app there are technical reasons for the separation of components: the frontend must integrate with a web server, and the backend must integrate with a database server; both of those servers must be configured correctly as well. However, should a researcher choose to use all three of our components together, this then implies running them all separately, which is less convenient than having a single combined module of some sort. One way around most of this complexity, that we plan to do in the future, is to build a configuration of `containers' to create an application stack (e.g. docker-compose\footnote{\url{docs.docker.com/compose/}} is popular software for managing multiple containers) with our components plus web server and database server configured within their own containers. However, even that option still requires access to a system installed with said container software and the necessary networking set-up and capabilities to serve web content on the Internet. If a researcher or citizen scientist does not have access to such a system, then it is possible to purchase access to one from a selection of many application server hosting providers.

\section{Conclusion}
We offer to the research community a free and open-source toolkit for downloading, processing and filtering VSVI from a given geographic region, and deploying a mobile-friendly human perception survey web app with a specially customized user interface on the resulting images. In the spirit of citizen science and the FAIR research principles, anyone may easily clone, modify and deploy this toolkit on any location of interest. We integrated it with the Mapillary platform because they provide open access to VSVI from many cities around the world, including some with little or no Google Street View coverage as of this writing such as Casablanca, Lusaka and Managua. We anticipate that this toolkit will be used to build deep learning models with the collected data and make predictions about human perception of large amounts of SVI over a wider area. Ultimately, we hope to see such perception data used in research that can help guide future planning and development choices, and therefore improve the quality of life for many people in urban areas.

\section*{Software \& data availability}
\begin{itemize}[noitemsep]
\item Name of software: Human perception and volunteered street view imagery project (percept)
\item Developer: Matthew Danish, m.r.danish@uu.nl
\item Source code / data: \url{github.com/Spatial-Data-Science-and-GEO-AI-Lab/percept}
\item Date first made available: February 2024
\item Hardware required: (server) Internet-connected server or virtual server; (client) Smartphone or desktop.
\item Software required: (server) Linux, Apache2, PostgreSQL, Node.js and Express.js; (client) Web browser.
\item License and cost: GNU General Public License 3.0; there is no cost.
\item Programming languages: JavaScript, HTML/CSS, Python
\end{itemize}
The available data has been processed to ensure it contains absolutely no personally-identifiable information and to prevent reconstruction of any such information, a problem which might occur in combination with other data sources.

\section*{Declaration of competing interest}
The authors declare that they have no known competing financial interests or personal relationships that could have appeared to influence the work reported in this paper.

\section*{Acknowledgements}

This work received funding (BMD 3.1fb220215) from the Faculty of Geosciences, Utrecht University, as well as from the FAIR Research IT program at Utrecht University. Neither of the funders had any role concerning the study design, data collection and analysis, interpretation, or dissemination. Maarten Zelymans Van Emmichoven helped with early conceptualization, and the Nederlands translation of the app.

\printcredits

\bibliographystyle{cas-model2-names}

\appendix
\appendixpage
\section{Survey details}
\subsection{Categories}\label{sec:categories}

Below are the categories used to gather ratings in the survey. Each participant was asked to rate 20 images in each category. The app presented five images consecutively under one category, and then the app randomly switched to another unexhausted category. For each category below is the corresponding category description, which was shown to each participant the first time they encountered the category. The same text was also subsequently available under a tooltip, for reference. The purpose of the category descriptions was not to be prescriptive but rather to help alleviate concerns about ambiguity or inspire participants who were uncertain about how to respond.

\begin{itemize}
    \item Walkability --
    \emph{Does this place look like an easy and safe place for people to travel on foot
or using a walking-equivalent mobility aid (e.g. wheelchair)?  This might
include factors such as the quality of sidewalks, pedestrian crossings, street
connectivity, and access to public amenities. Walkable communities encourage
people to walk or use other non-motorized modes of transportation.}
    \item Bikeability --
    \emph{Does this place look accessible, attractive, safe and convenient for cycling as a mode of general-purpose transportation, or cycling-equivalent mobility aid
(e.g.  mobility scooter)? This might include factors such as cycle lanes,
tracks, and parking, as well as the overall design of streets, junctions and
any visible surroundings.}
    \item Pleasantness --
    \emph{Does this place look enjoyable or pleasing to the senses or emotions? This
might include factors such as the aesthetics of the surroundings, the quality
of the air and lighting, the soundscape, and the presence of other people or
natural elements.}
    \item Greenness --
    \emph{Rate the apparent amount of vegetation and greenery in a given environment.
This encompasses the presence of trees, shrubs, plants, and other natural
elements.}
    \item Safety --
    \emph{Does this place look like you would feel protected from harm or danger, in
terms of personal safety and security? Do you believe that it is likely that
you would feel safe here at all times of day or night? This might include the
presence of elements such as lighting, good maintenance, presence of other
people and natural surveillance.}
\end{itemize}

\subsection{Socio-demographic information}\label{sec:demographics}

\begin{itemize}[noitemsep]
    \item Age
    \item Gender
    \item Estimated monthly income
    \item Level of education
    \item Home postal code and country of residence
    \item Data usage consent
\end{itemize}

We gather several pieces of socio-demographic information  to make comparisons between responses from users of different backgrounds.
We strove to keep the number of demographic questions limited to five because we did not want to create an unnecessarily high barrier to entry. For example, it would be helpful to know each participant's country of birth and upbringing, as well as country of residence, but we felt that would be too confusing.
Of the information collected, only age and data consent are required for participation in the survey. Age is required because we want the users to self-certify that they are 18 or older, to avoid complications with obtaining data usage consent from minors.

The other questions are optional and relatively free-form, with the understanding that we may not be able to interpret all answers given. `Level of education' is the most structured question, with four possible options (`Primary', `Secondary', `Tertiary' and `Postgraduate'). This is a compromise between legibility and specificity because, for example, in the Netherlands it is quite common to break down education level into categories like MBO, HBO, and VWO, but these abbreviations have no meaning to people who are not familiar with the Dutch system.

\section{Backend details}

\subsection{Public API v1}\label{sec:publicapi}

The backend server defines a URL for each API function, of the form \texttt{/api/v1/<function>} where \emph{<function>} is one of the following, along with corresponding form parameters:

\begin{itemize}[noitemsep]
\item \apifn{newperson} (\apiparam{age}, \apiparam{monthly\_gross\_income}, \apiparam{education}, \apiparam{gender}, \apiparam{country}, \apiparam{postcode}, \apiparam{consent})
\begin{itemize}[noitemsep]
\item Creates a new participant based on the given socio-demographic inputs (only \apiparam{age} and \apiparam{consent} are required).
\item Returns as a JSON dict \apiparam{session\_id} and \apiparam{cookie\_hash}, the latter of which is intended to be stored as a cookie in the participant's browser if they consent to the study.
\end{itemize}
\item \apifn{getsession} (\apiparam{session\_id} or \apiparam{cookie\_hash}; either can be used, depending upon what is known)
\begin{itemize}[noitemsep]
\item Finds the corresponding \apiparam{session\_id} for a given \apiparam{cookie\_hash}, or vice versa.
\item Returns as a JSON dict \apiparam{session\_id} and \apiparam{cookie\_hash} both fully filled out.
\end{itemize}
\item \apifn{fetch} (\apiparam{session\_id})
\begin{itemize}[noitemsep]
\item Select an image from the database that has not yet been rated by the current participant.
\item Returns as a JSON dict \apiparam{cityname}, \apiparam{url} and \apiparam{image\_id}
\end{itemize}
\item \apifn{new} (\apiparam{session\_id}, \apiparam{cookie\_hash}, \apiparam{image\_id}, \apiparam{category\_id}, \apiparam{rating})
\begin{itemize}[noitemsep]
\item Creates a new rating data point for the current session.
\item Returns the same result as \apifn{countratingsbycategory}.
\end{itemize}
\item \apifn{undo} (\apiparam{session\_id}, \apiparam{cookie\_hash})
\begin{itemize}[noitemsep]
\item If permitted (see Appendix~\ref{sec:undoprotocol}) then undo the most recent rating by the participant.
\item Returns the same result as \apifn{countratingsbycategory}.
\end{itemize}
\item \apifn{countratingsbycategory} (\apiparam{session\_id})
\begin{itemize}[noitemsep]
\item Returns a JSON dict with a single element, \apiparam{category\_counts}, which in turn contains a JSON dict keyed by category ID with information about how many images have been rated by the current participant in each corresponding category. For example, \apiparam{result[`category\_counts'][1]} gives the number of ratings that have been submitted for the category with ID 1.
\end{itemize}
\end{itemize}

\subsection{Undo protocol}\label{sec:undoprotocol}

The backend enforces a particular undo protocol: only the single most recent rating can be undone. This ensures that the use of undo is limited to only correcting a simple mistake and does not result in large-scale deletion of data from the database. Skipped images are not reported to the backend at all; undo of skips is possible and it is handled entirely in the frontend, in a transparent manner so that the participant cannot tell the difference between undoing a skip or undoing a rating.

\section{Processing and filtering VSVI details}

For full up-to-date usage information please see the Spatial Data Science and Geo AI Lab\footnote{\url{github.com/Spatial-Data-Science-and-GEO-AI-Lab/percept}} web site.

\subsection{Downloading imagery from Mapillary}\label{sec:mapillarydl}

\noindent\filename{mapillary\_jpg\_download.py}

    This script takes a Mapillary API key\footnote{\url{www.mapillary.com/developer}} and a bounding box (west, south, east, north) and conducts a lengthy but robust and restartable procedure to methodologically find and download each tile data file, cache it, and then download all of the eligible SVI that is found within the tile (and within the bounding box).

    To give more detail: each tile data file is a set of GeoJSON 
    features from within a certain pre-defined rectangular area. In our case, we are interested only in features corresponding to SVI. Therefore the tile files are lists of images (photographs) with the following pieces of information for each one: a unique image identifier, sequence identifier corresponding to a series of photographs taken in a row (often while driving down a street), the compass angle at which this photograph was taken, geographic latitude and longitude coordinates, the time it was taken, and a Boolean value indicating whether or not it is a panoramic photograph.

    With the image identifier, we are able to use the Mapillary API to obtain the precise URL of the original photographic image, and then download it. The photographs are organised by sequence identifier (for later reference) and stored in filenames corresponding to the image identifier (which is unique). In this way, we have access to all of the available information about each photograph: the imagery is stored in a directory structure organised by sequence identifier, and the unique image identifying number can be used to look up all of the meta-information in the tile data files.

    Should any part of the downloading process fail, the script automatically backs up and restarts the download process, with exponential back-off up to a limited number of retries. After that point, if the download still fails, the image identifier can be (optionally) saved to a file containing a list of failed identifiers. Later, the process can be restarted using the cached tile data files, and the script can be directed to focus on the failed image identifiers (or any image identifiers the user chooses). In our experience, the downloading process does fail from time to time, and therefore this functionality was very valuable, as the full download process can take days depending on how large the required region is.

\subsection{Semantic segmentation of imagery}

\noindent\filename{torch\_segm\_images.py}

    This script is designed to work with very large directories or lists of filenames corresponding to SVI. It uses the PyTorch library and by default the {\tt facebook/mask2former-swin-large-cityscapes-semantic} model~\citep{cheng2022} for image segmentation. The result is a matrix with values corresponding to the meaning of each pixel in the input image. These matrices are then stored in compressed numpy array files (.npz) alongside each image. This step takes a substantial amount of time but thanks to the PyTorch library it can be significantly sped-up with the assistance of a GPU.

\subsection{Cropping and filtering segmented imagery}\label{sec:processsegm}
\noindent\filename{torch\_process\_segm.py}

    This script uses the previously obtained segmentation arrays and applies the remaining processing to crop panoramic images and determine which non-panoramic images should be discarded. The results of this script are a series of subimages from each panoramic image, saved alongside it, and a series of output logs for each image, detailing the analysis results and findings for each image. These findings can be used to select images for acceptance or rejection based on image quality. The script also outputs SQL statements for insertion into the percept-backend database (initially in a disabled state), and further statements for enabling the use of the images when ready to show to end-users.

\subsection{Finding road center-lines}\label{sec:roadcenters}

\begin{figure*}[tb!]\centering
    \subfigure[][\label{fig:panomaskstagesB}Panoramic photograph with semantic segmentation highlighting: road in black, and non-road in gray.]{\includegraphics[width=0.8\textwidth]{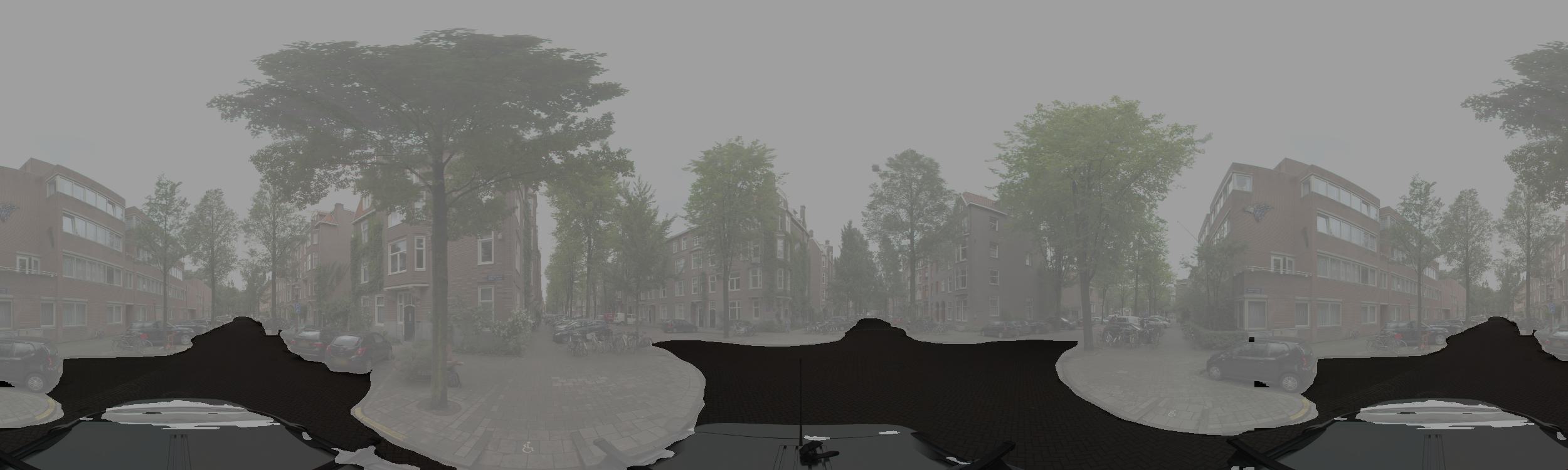}}
    \subfigure[][\label{fig:panomaskstagesC}Vertical red lines show the estimated road center-lines according to the classic Hough transform method.]{\includegraphics[width=0.8\textwidth]{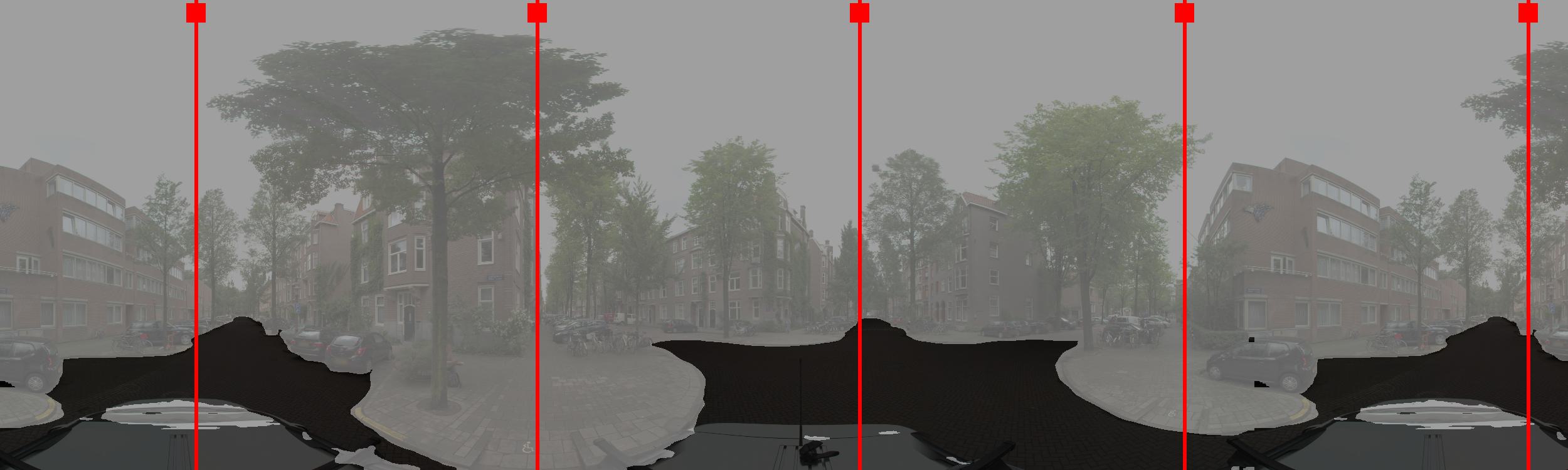}}
    \subfigure[][\label{fig:panomaskstagesD}Vertical green lines show the estimated road center-lines according to our segmentation-based method.]{\includegraphics[width=0.8\textwidth]{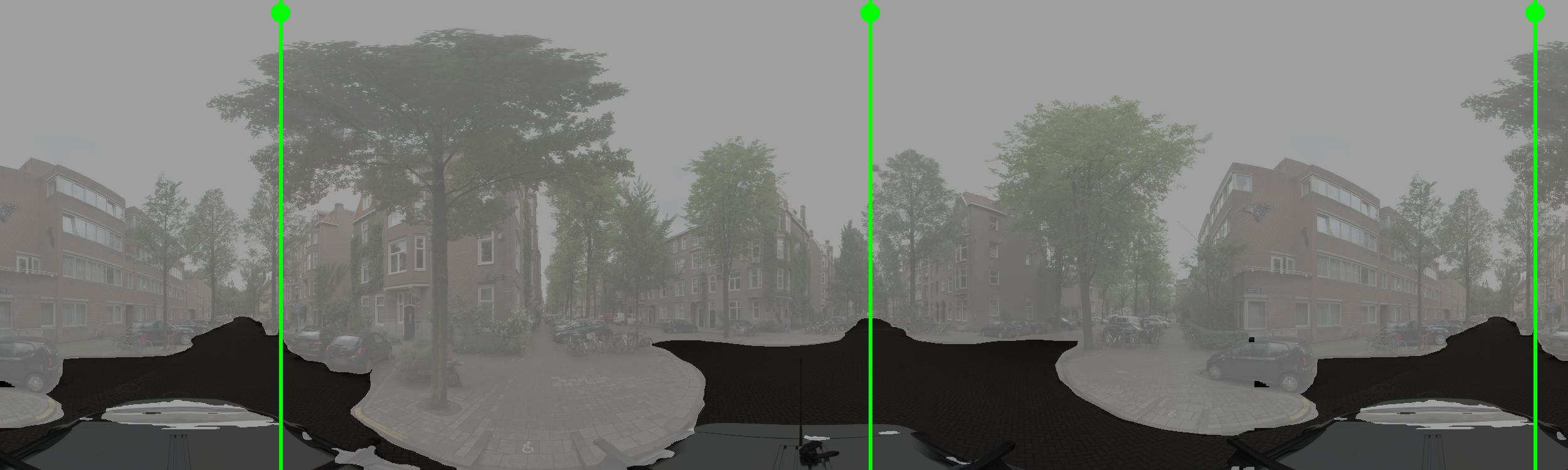}}
    \subfigure[][\label{fig:panomask}The original panoramic photograph overlaid with both kinds of estimated road center-line detection method results. The red lines (with squares) come from the Hough transform method, and the green lines (with circles) come from our segmentation-based method.]{\includegraphics[width=0.8\textwidth]{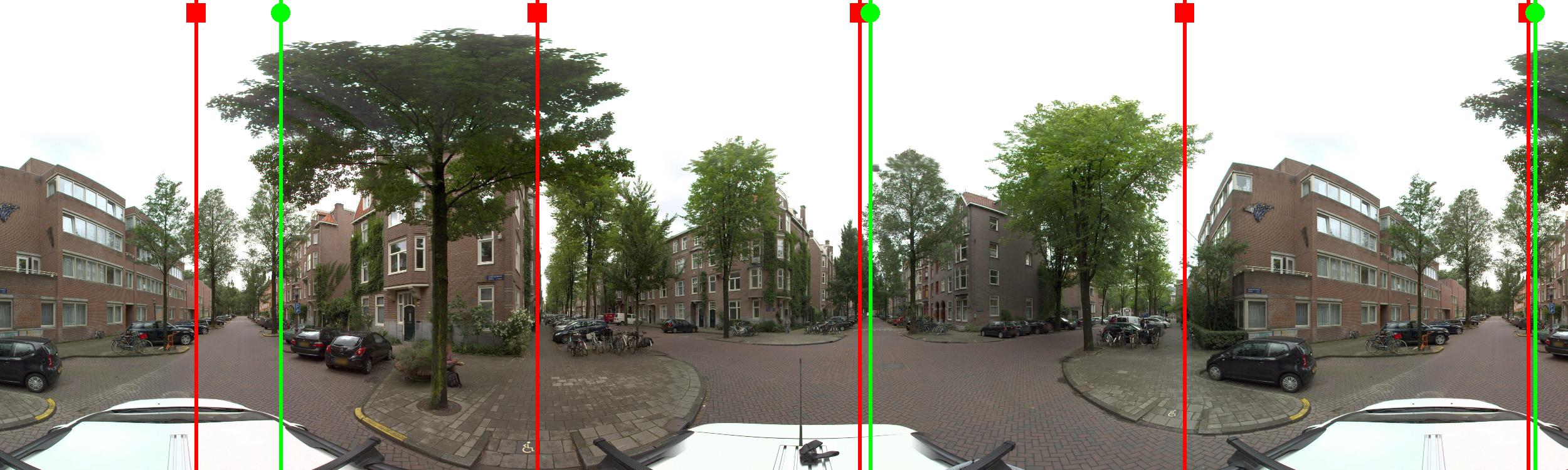}}
    \caption{An example panorama, showing semantic segmentation and two methods of road center-line finding.}
    \label{fig:panomaskstages}
\end{figure*}

We show the intermediate stages of road center-line finding in Figure~\ref{fig:panomaskstages}. Before that, the zeroth step (not shown) is simply to crop out the bottom quarter of the panoramic photograph because it usually only shows the vehicle holding the camera, and then to extend the panorama by wrapping the first 25\% of the photo onto the right-hand side of the photo. The result will be that the first fifth of the extended photo is exactly the same as the final fifth, which can be easily seen in the given example. The reason for doing this is to prevent analysis from missing out on any boundary cases on the left or right edge of the panoramic photograph; instead we have to deal with duplicate results, which we clean up later.

We then apply semantic segmentation \citep{thisanke2023},
in this case using the Mask2Former model \citep{cheng2022} trained on the CityScapes dataset~\citep{cordts2016CVPR}, readily available through the HuggingFace transformers library~\citep{wolf2020}. We chose this off-the-shelf model for no reason other than because it was the most popular and relevant semantic segmentation model available for rapid deployment from Huggingface. We found that it works sufficiently well for the roads we studied in Amsterdam but we do not depend on any particular property of it. Therefore, users preparing their own imagery with our toolkit could substitute another model, if desired. We are only interested in finding the portion of the image that corresponds to `road', therefore Figure~\ref{fig:panomaskstagesB} shows the road pixels highlighted in black, and everything else is grayed out.

One of the classic road center-line finding algorithms \citep{bhasker2022}
tries to find vanishing points in photographs, with the idea that roads tend to be linear features following perspective lines into the distance. To do this, we simply apply the venerable Hough transform~\citep{illingworth1988} as available in OpenCV~\citep{opencv_library}, to edges in an image and then finds the points of intersection. In Figure~\ref{fig:panomaskstagesC} we show the results of such an algorithm that has been applied to edges found (by the OpenCV Canny edge detection) in the semantic segmentation matrix, as indicated by the red lines (and tagged with squares). You can see that it finds some roads, but it also finds foot-ways. It also gets confused when near the edge of the photograph. The first and the last red lines should appear in the same relative position (to the road) because these portions of the photo are the duplicated (wrapped) first and final fifths of the photo, but in fact the lines are placed differently. This happens because some perspective lines in the first fifth, coming from the central portion of the panoramic image, do not exist in the final fifth.

\paragraph{Our segmentation-based method}

We chose a more focused method of finding center-lines. We use semantic segmentation on images to identify the pixels corresponding to roads. For each column $x$ in the segmentation matrix, let \B x be distance between the bottom of the matrix and topmost `road'-labeled pixel, and let \C x be the count of `road'-labeled pixels in the bottom half of the matrix. Choosing an adjustment factor $k$ (in our case, $k=1/8$) then we define \R x as a combination of the above two: $\R x = \B x + k\C x$. While \B x captures the intuition that roads should appear as `peaks' in the segmentation matrix, there are sometimes spurious road pixels. Therefore, \C x ensures that there is a substantial number of road pixels behind each peak. However, \C x by itself has the problem that the camera-carrying vehicle often interferes with the segmentation results near the bottom of the image, especially around the centers of roads. Therefore we apply a scaling factor $k$ to ensure that the sides of roads are not overemphasized in the output.

Using the Python Scipy library~\citep{virtanen2020scipy}, we then find the peaks of \R x from left-to-right across the width of the image, under the assumption that there is one valid road to be found approximately in each third of the panorama. This relatively simple algorithm is surprisingly effective, and we found it to be more effective than the above road-finding algorithms using vanishing point perspective detection. An example is shown by the green lines (tagged with circles) in Figure~\ref{fig:panomaskstagesD}.

Most notably, this method finds the centers of roads where the camera is looking directly down the road. This matches the intuitive description of what we are seeking: the view that person has when they stand in a road and look down it. Further notes: the first and third line in the example are in the same relative position, as they should be, since these sections of the image are duplicates. In both cases, the estimated center-line is a bit off from where it should be; this has been caused by spurious `non-road' pixels from the intrusion of the camera-carrying vehicle into the image.

Figure~\ref{fig:panomask} shows both methods overlaid onto the original panorama. In our experience, the segmentation-based method generally gave the best estimates, and even if it was off, it was not by much and the end result remained presentable. It would be better if the camera-carrying vehicle could be cut out entirely, however it does not appear in a consistent way nor is it detected consistently by semantic segmentation. Therefore, we applied a pragmatic rule of thumb and cropped the bottom quarter of the panorama, seeing that the horizon on these panoramic images always falls within the middle band to a reasonable extent.

\subsection{Image quality filtering}\label{sec:qualitythreshold}

The script \filename{torch\_process\_segm.py} (see Appendix~\ref{sec:processsegm}) performs a number of operations for each given image, which includes computing the two key factors of our image quality algorithm: (1) `contrast' ($C$) as calculated by a function derived from the Scikit-image function \texttt{skimage\_contrast}~\citep{van2014scikit}, and (2) `tone-mapping score' ($T$) as described by \citet{stefanescu2021imagequality}, which effectively tries to measure if an image is too dark, too bright or suffers from a poorly distributed range of colors as shown in a color-histogram breakdown of the image. Both $C$ and $T$ are numbers between 0 and 1. We define $C_{min}=0.35$ and $T_{min}=0.35$ as thresholds for the following tests, and $T_{floor}=0.8$ as a threshold for an adjustment factor. Both tests must be satisfied for the image to pass our quality filter:

\begin{flalign*}
  T_{min} &< T \\
  C_{min} &< C + \text{max}(0, T - T_{floor})
\end{flalign*}

These tests effectively establish minimum contrast and tone-mapping score requirements for images. The threshold numbers were chosen by trial and error. We also adjusted the contrast test so that a particularly high tone-mapping score could compensate for a worse contrast result. We found several cases where images had slightly lower contrast than desired but had strong tone-mapping scores and they appeared reasonable to the eye. Therefore, we incorporated this adjustment factor into the contrast test so that we would not be overly pessimistic and lose those images.

\end{document}